\documentclass[letterpaper, 10 pt, conference]{ieeeconf} 
\usepackage{amsmath,amsfonts}
\usepackage{algorithmic}
\usepackage{algorithm}
\usepackage{array}
\usepackage[caption=false,font=normalsize,labelfont=sf,textfont=sf]{subfig}
\usepackage{textcomp}
\usepackage{url}
\usepackage{verbatim}
\usepackage{graphicx}
\usepackage{cite}
\usepackage{nomencl}
\usepackage{ntheorem}   
\usepackage{lipsum}     
\usepackage{graphicx}

\usepackage{xcolor}
\usepackage{hyperref}

\hypersetup{
	colorlinks=true,
	linkcolor=black,
	filecolor=black, 
	urlcolor=black,
	citecolor=black
}

\usepackage{bm}

\DeclareRobustCommand{\uvec}[1]{{%
		\ifcsname uvec#1\endcsname
		\csname uvec#1\endcsname
		\else
		\bm{\mathbf{#1}}%
		\fi
}}

\IEEEoverridecommandlockouts                              

\usepackage{graphicx} 

\title{\LARGE \bf
PenduMorph: Development and Motion Analysis of Pendulum-Actuated Rolling Reconfigurable Spherical Robot with Magnetic-Coupling
}

\author{Aung Myat*, Peter Noyce*, May Forgan*, Qing Yu* and Seyed Amir Tafrishi
\thanks{Authors are with Geometric Mechanics and Mechatronics in Robotics (gm$^2$R) Lab, the School of Engineering, Cardiff University, Cardiff, CF24 3AA, United Kingdom}
\thanks{* These authors contributed equally. }
\thanks{Seyed Amir Tafrishi is the corresponding author of this study (phone: +44 29208 76176, e-mail: {\tt\small Tafrishisa@cardiff.ac.uk}).}}

\renewcommand{\baselinestretch}{.89}
\begin{document}

\maketitle
\thispagestyle{empty}
\pagestyle{empty}


\begin{abstract}
This paper presents "PenduMorph", a wireless reconfigurable rolling spherical robot designed as a modular platform for enclosed locomotion and inter-module interaction in challenging environments. The proposed robot extends our previous pendulum-actuated rolling disk concept to a fully enclosed spherical architecture integrating a 2-DoF internal pendulum, onboard control, battery-powered operation, and magnetic docking. The design aims to combine independent rolling mobility with protected hardware and reliable reconfigurability. We first present the robot design and an analytical study of the magnetic coupling mechanism to evaluate retention and interaction between coupled modules. We then experimentally investigate key motion behaviors at both the single-module and dual-module levels, including independent rolling, magnetic coupling, and coordinated coupled motion. The results show that the proposed platform enables stable wireless operation and a set of distinctive reconfigurable rolling behaviors, providing a useful foundation for future modular spherical robots operating in contact-rich and demanding environments.
\end{abstract}

\section{INTRODUCTION}

Reconfigurable robots derive their versatility from the ability to alter morphology, connectivity, and functional role across a team of modules. This capability is particularly attractive for inspection, search, and other challenging tasks, where the robot may need to adapt its geometry to confined spaces, obstacles, or changing operational demands~\cite{981854,moubarak2012modular,seo2019modular}. However, many modular systems still rely on exposed docking hardware, external appendages, or restricted locomotion modes, which can limit robustness in practice. For modular reconfigurable platforms to become more useful in harsh environments, each module should ideally retain independent mobility, self-contained actuation, and protected hardware while still being able to dock, undock, and cooperate reliably with neighbouring modules.

Rolling robots provide an appealing route toward this objective because locomotion can be generated through internal actuation while the outer body remains compact and mechanically enclosed~\cite{armour2006rolling,tafrishi2019design}. Such closed-body rolling platforms are particularly attractive in dusty, cluttered, or contact-rich environments, where impact tolerance and hardware protection are important. Prior spherical rolling robots have demonstrated the potential of internally actuated shells, from early rotor-driven concepts to motion-planning formulations for internally driven spheres~\cite{bhattacharya2000spherical,morinaga2014motion}. More recent platforms have further emphasized manoeuvrability and payload accommodation within enclosed spherical bodies~\cite{belzile2022aries}. Reconfigurable rolling systems have also been explored in reaction-torque or cart-based forms~\cite{liang2020freebot}; however, such approaches can suffer from sticking, reduced mobility in certain configurations, and limited suitability for fully enclosed sensing and actuation layouts. These limitations motivate pendulum-driven rolling mechanisms, which can shift the centre of mass more directly, improve stability relative to cart-based layouts, and offer a more natural route to multi-DoF internal actuation. Nevertheless, rolling platforms that simultaneously combine enclosed morphology, sufficiently strong internal actuation, and reliable inter-module reconfigurability remain relatively underdeveloped.

\begin{figure}[t!]
      \centering
      \includegraphics[width = 0.5\textwidth]{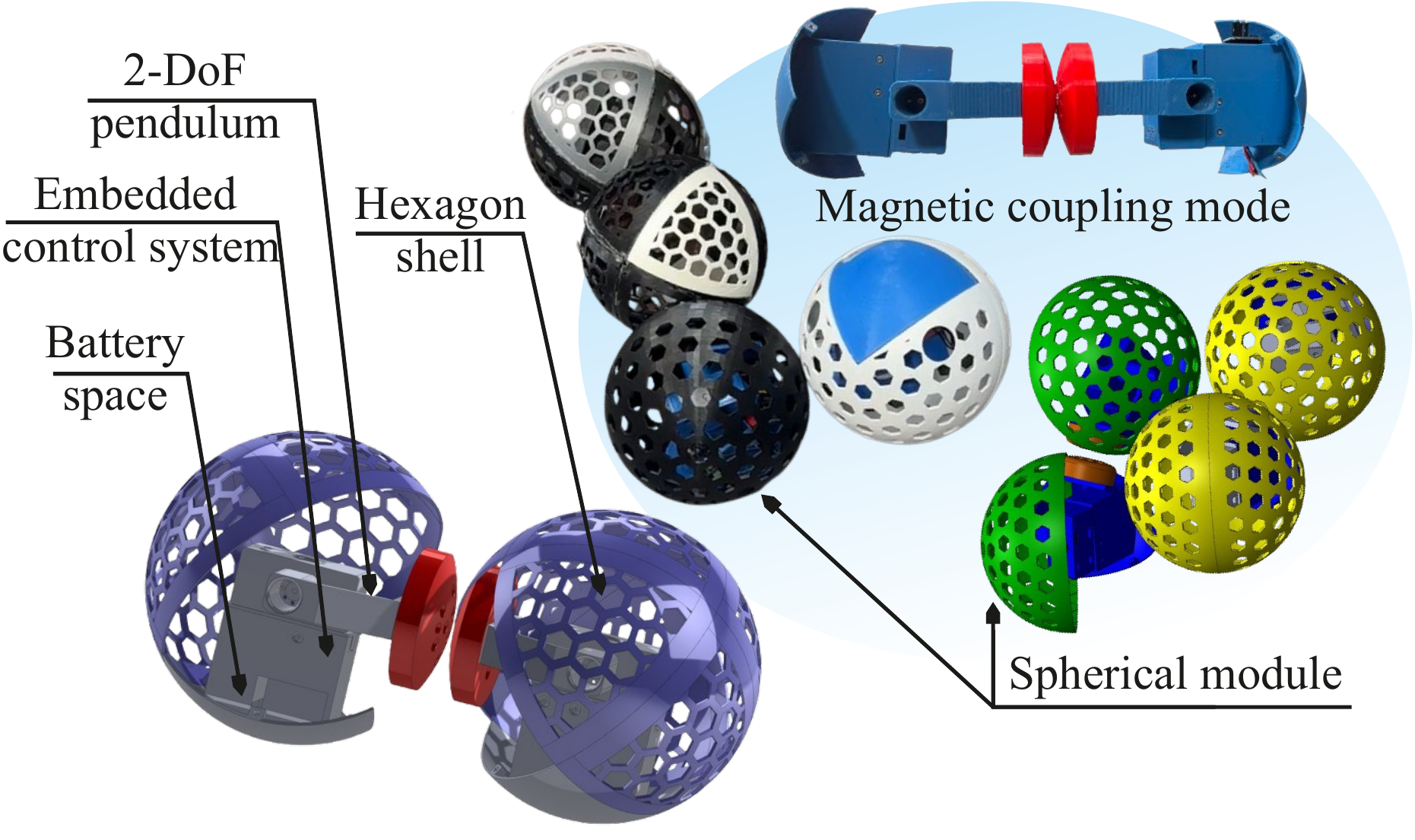}
      \caption{Reconfigurable rolling spherical robot design with magnetic coupling.}
      \label{fig:robot_design}
\end{figure}

Modular reconfigurable robots consist of individual modules with local sensing, actuation, and control that can operate independently or as part of a larger assembly~\cite{seo2019modular}. Depending on the morphology and connection architecture, modules may dock through mechanical interfaces, magnetic coupling, gripping systems, or hybrid combinations, and may support one-to-one or one-to-many connectivity~\cite{moubarak2012modular,seo2019modular,liang2023decodingmodularreconfigurablerobots}. For rolling modular robots, the main difficulty is not only forming a reliable connection, but maintaining controllable motion once contact, slip, and rolling constraints interact. This is closely related to non-prehensile manipulation, where motion depends strongly on friction, contact evolution, and dynamic feasibility~\cite{ruggiero2018nonprehensile}. As also highlighted in recent rolling-contact surveys~\cite{tafrishi2025survey}, these effects remain a central obstacle to predictable and scalable behaviour in modular rolling systems.

Our recent work began to address this problem through a pendulum-actuated reconfigurable rolling disk robot with permanent-magnet coupling~\cite{Ollie2024towards}. That disk-based platform serves as the immediate predecessor of the present spherical design and motivates the transition to a fully enclosed rolling body with more uniform contact geometry and we studied potential rolling-slipping behavior in analytical simulation as novel nonprehensile manipulation problem \cite{wiltshire2025novel}. That study demonstrated that internal pendulum actuation, combined with passive magnetic docking, can produce useful reconfigurable behaviours including independent rolling, inter-module coupling, relative rotation, and preliminary balancing. However, the disk geometry also imposes clear limitations: contact is directionally biased, full-body coverage is absent, and the exposed structure is less suitable for physically demanding environments. These limitations motivate the transition from a rolling disk to a rolling spherical robot, where the shell can provide uniform contact, improved protection, and a more natural platform for enclosed modular rolling interaction.

In this paper, we present \emph{PenduMorph}, a wireless reconfigurable spherical rolling module that extends our previous pendulum-driven magnetic-coupling concept into a fully enclosed platform with onboard actuation, sensing, control, and docking capability. First, the mechanical design of the spherical module and its embedded wireless architecture are introduced in Section~\ref{Sec:robordesign}. Next, an analytical model of the magnetic coupling mechanism is developed to characterize inter-module interaction and coupling retention. The experimental implementation, communication architecture, trajectory generation, and motion reconstruction framework are then presented in Section~\ref{sec:exp_control}. Finally, the motion capabilities of the platform are investigated experimentally at both the single-module and coupled-module levels, including independent rolling, coupling, and coordinated interaction, with the corresponding results reported in Section~\ref{sec:results}. The main contributions of this work are threefold: 1) the development of a wireless pendulum-actuated spherical rolling module with integrated magnetic docking; 2) an analytical characterization of the magnetic coupling interface; and 3) an experimental investigation of independent rolling, coupling, and coupled-motion behaviours in a reconfigurable spherical robotic platform.


\section{RECONFIGURABLE ROLLING ROBOT}\label{Sec:robordesign}
\begin{figure}[t!]
      \centering
      \includegraphics[width = 0.4\textwidth]{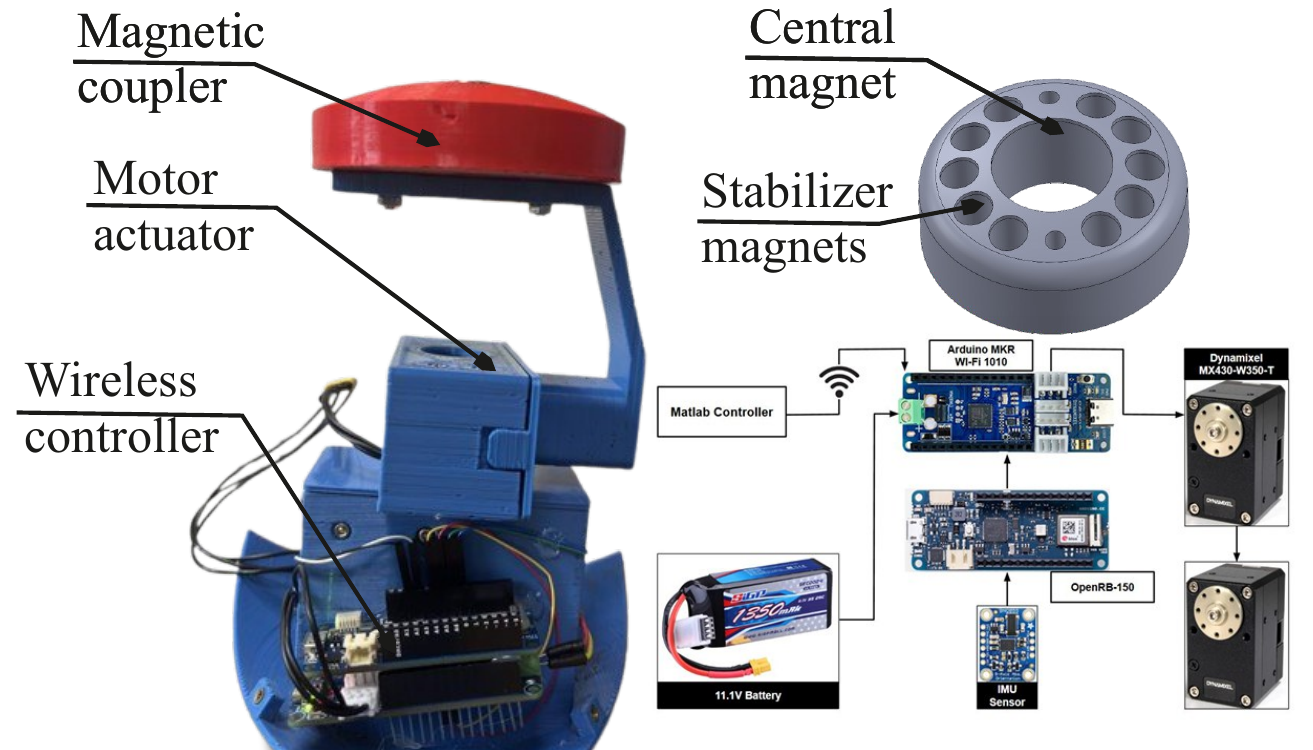}
      \caption{Mechanical and embedded architecture of each spherical module, showing the pendulum actuator, magnetic docking unit, and wireless control/sensing pipeline.}
      \label{fig:Circuit_Design}
\end{figure}

This section introduces the mechanical design, embedded architecture, and magnetic docking interface of the reconfigurable spherical robot. Each module is realized as a self-contained rolling unit, comprising a spherical body, an internally actuated pendulum mechanism, onboard sensing, and a magnetic docking interface, as shown in Fig.~\ref{fig:Circuit_Design}. The locomotion principle is based on a 2-DoF pendulum-driven barycentric actuation mechanism, in which two Dynamixel servo motors drive the internal mass to generate rolling motion by shifting the centre of mass of the sphere. The pendulum apex also serves as the docking end-effector, integrating the magnetic coupler so that the same internal mechanism supports both locomotion and inter-module interaction. In this work, two active modules are used to study independent rolling, docking, coupled motion, and the resulting interaction behaviours.

The embedded system is designed for fully wireless operation. A host MATLAB controller communicates with each robot through WiFi using an Arduino MKR WiFi 1010, which serves as the wireless communication interface. The MKR exchanges commands and state data with an OpenRB-150 controller through a UART serial link, while the OpenRB-150 directly actuates the two Dynamixel motors through the Dynamixel TTL bus. A BNO055 inertial measurement unit is connected via I\textsuperscript{2}C to provide attitude and angular-rate feedback, and the entire onboard system is powered by an 11.1~V battery pack. This layered architecture separates high-level wireless supervision from low-level motor execution, enabling untethered operation and repeatable experimental evaluation.

The magnetic docking unit consists of a central magnet surrounded by an array of stabilizer magnets. The central magnet provides the primary attractive force required to maintain coupling between neighbouring modules, acting as the main load-bearing magnetic interface during contact. The surrounding stabilizer magnets improve passive alignment during approach, reduce off-axis tilting and rotational mismatch after contact, and help suppress local vibration or chatter at the interface once coupling is established. As a result, the interaction becomes smoother and more repeatable during both coupling and decoupling, while also improving robustness to small pose errors during docking. This combined magnetic layout therefore supports not only attachment strength, but also passive self-alignment and interaction stability, which motivates the analytical treatment of the magnetic coupling mechanism presented next.

The magnetic coupling between the two pendulum caps is modeled as the interaction of two parallel arrays of axially magnetized cylindrical NdFeB magnets. Each cap contains one central magnet and \(N_r\) surrounding magnets placed on a pitch circle of radius \(r_m\). Since plastic PLA is nonmagnetic to a good approximation, the two printed walls are treated as part of the effective nonmagnetic stand-off. Thus, the total face-to-face separation is
\begin{equation}
g = 2h_{\mathrm{PLA}} + g_a ,
\label{eq:mag_gap}
\end{equation}
where \(h_{\mathrm{PLA}}\) is the thickness of one shell wall and \(g_a\) is the remaining air gap.

The in-plane magnet-center locations are written as
\begin{align}
\mathbf{r}_0 &= \mathbf{0}, \\
\mathbf{r}_k &= r_m
\begin{bmatrix}
\cos\theta_k\\
\sin\theta_k
\end{bmatrix},
\qquad
\theta_k=\frac{2\pi(k-1)}{N_r},
\quad
k=1,\dots,N_r .
\label{eq:ring_positions}
\end{align}
The central magnet is assigned a reference facing polarity, while the surrounding magnets alternate in sign around the ring. Hence,
\begin{equation}
\lambda_0 = 1,
\qquad
\lambda_k = (-1)^{k-1},
\qquad
k=1,\dots,N_r ,
\label{eq:ring_polarity}
\end{equation}
and the opposing cap is assembled with opposite facing polarity at each corresponding location, i.e., \(\bar{\lambda}_k=-\lambda_k\).

Following the uniformly magnetized cylindrical-magnet formulation, the axial interaction between magnet \(i\) on the upper cap and magnet \(j\) on the lower cap is obtained from the four pole-face interactions as
\begin{equation}
F_{ij}
=
\lambda_i \bar{\lambda}_j
\frac{B_{r,i} B_{r,j}}{4\pi\mu_0}
\sum_{\alpha=0}^{1}\sum_{\beta=0}^{1}
(-1)^{\alpha+\beta}
\mathcal{I}_{ij}(d_{ij},z_{\alpha\beta}),
\label{eq:Fij}
\end{equation}
where \(\mu_0\) is the permeability of free space, \(d_{ij}=\|\mathbf{d}_{ij}\|\) is the lateral offset,
\begin{equation}
\mathbf{d}_{ij}=\mathbf{r}_i-\mathbf{r}_j ,
\label{eq:dij_vec}
\end{equation}
and
\begin{equation}
z_{\alpha\beta}=g+\alpha h_i+\beta h_j .
\label{eq:zab}
\end{equation}
Here, \(h_i\) and \(h_j\) denote the thicknesses of magnets \(i\) and \(j\), respectively. The geometric kernel \(\mathcal{I}_{ij}\) is defined by the continuous surface integral \cite{vokoun2009magnetostatic}
\begin{equation}
\mathcal{I}_{ij}(d,z)
=
\iint_{A_i}\iint_{A_j}
\frac{z}
{\left(\|\boldsymbol{\xi}-\boldsymbol{\eta}-\mathbf{d}_{ij}\|^2+z^2\right)^{3/2}}
\,\mathrm{d}A_{\eta}\,\mathrm{d}A_{\xi},
\label{eq:Iij}
\end{equation}
where \(A_i\) and \(A_j\) denote the circular pole-face areas of the two magnets. Eq.~\eqref{eq:Iij} is the continuous magnetostatic term, while \eqref{eq:Fij} compactly accounts for the finite cylinder thickness through the four face combinations. Next, the total cap-to-cap axial force is then
\begin{equation}
F_m=\sum_{i=0}^{N_r}\sum_{j=0}^{N_r}F_{ij}.
\label{eq:Fm_general}
\end{equation}
For the present alternating ring, the net central-to-ring contribution cancels by symmetry because \(\sum_{k=1}^{N_r}\lambda_k=0\). Therefore, the result can be written more compactly as
\begin{equation}
F_m = F_c + N_r\sum_{\ell=0}^{N_r-1}\nu_{\ell} F_s(d_{\ell}),
\label{eq:Fm_reduced}
\end{equation}
where \(F_c\) is the central-pair contribution, \(F_s(d_{\ell})\) is the contribution associated with ring chord class \(d_{\ell}\), and
\begin{equation}
d_{\ell} = 2r_m\sin\!\left(\frac{\ell\pi}{N_r}\right),
\qquad
\nu_{\ell}=-(-1)^{\ell} .
\label{eq:dm}
\end{equation}

For the prototype parameters listed in Table~\ref{tab:prototype_params}, the cap-to-cap attraction predicted by the reduced force model in \eqref{eq:Fm_reduced}, with the pairwise interaction defined through \eqref{eq:Fij}--\eqref{eq:Iij} and the ring chord distances given by \eqref{eq:dm}, yields a central-pair contribution of \(F_c\approx114.7\,\mathrm{N}\). The directly opposed surrounding magnets provide a further \(47.2\,\mathrm{N}\); however, the finite pitch-circle geometry and alternating polarity distribution encoded in \eqref{eq:Fm_reduced} introduce off-axis ring-to-ring interactions that reduce this gain by approximately \(16.9\,\mathrm{N}\). Consequently, the net axial coupling force is \(F_m\approx145.0\,\mathrm{N}\), which corresponds to an equivalent static holding mass of approximately \(14.8\,\mathrm{kg}\). These results indicate that the proposed pendulum-cap coupling is theoretically sufficient for nominal axial retention, while the experiments reported later provide functional validation of repeated docking and retained coupled interaction. In particular, the central magnet acts as the primary load-bearing element, while the surrounding ring mainly enhances passive centering and rotational registration, and still contributes a useful net increase in retention despite the geometric cancellation associated with the alternating polarity layout \cite{vokoun2009magnetostatic,shinetsu2018n52}.

Although \eqref{eq:Fm_reduced} is quasi-static, it still provides a useful first criterion for coupling integrity under pendulum motion. Dynamic rotation does not directly weaken the magnetostatic attraction, but it can generate a normal opening force \(F_n\) and a peeling moment \(M_p\) at the interface. A compact retention condition is
\begin{equation*}
F_n + \frac{|M_p|}{r_m} < F_m ,
\end{equation*}
so that, for the present prototype, decoupling is expected only when
$
F_n + |M_p|/r_m > 145.0\,\mathrm{N}.
$
This indicates that the proposed cap coupling retains substantial axial margin, although dynamic tilt or impact can still reduce the effective holding capacity by introducing local gap growth and peel-type loading. A direct force-gauge validation of the analytical model is left for future work; in the present study, the model is instead supported indirectly through repeated successful docking and retained coupled motion in the experiments.

\section{Experimental Implementation and Motion Reconstruction}
\label{sec:exp_control}

This section summarizes the experimental framework used to operate the modular rolling robot and analyse its measured motion. It first introduces the supervisory communication architecture linking the host controller to the embedded electronics of each module. It then presents the smooth trajectory generation and joint-space tracking law used during experiments. Finally, it describes the post-processing method used to reconstruct planar rolling motion from the IMU measurements.

\subsection{Supervisory Communication Architecture}

The modular robot is operated through a centralized supervisory architecture in which a host PC acts as the master controller and communicates with the embedded electronics of each module over TCP/IP. Within this framework, the PC performs reference generation, state acquisition, command dispatch, and data logging, while low-level execution is carried out onboard by the embedded controller. This arrangement is convenient for coordinated experiments because it separates high-level supervision from low-level motor communication.

To keep the formulation independent of implementation-specific firmware details, the communication layer is represented at the \(k\)-th sample by
\begin{equation}
\boldsymbol{\pi}_k=
\begin{bmatrix}
\mathbf{h}^{\top} & m_k & s_k & \mathbf{u}_k^{\top} & \chi_k
\end{bmatrix}^{\top},
\label{eq:packet_general}
\end{equation}
where \(\mathbf{h}\) is a fixed synchronization header, \(m_k\) is the command mode, \(s_k\) is the packet sequence index, \(\mathbf{u}_k\in\mathbb{R}^{n}\) is the command vector transmitted from the master PC, and \(\chi_k\) is the packet checksum. Here, \(n=\sum_{i=1}^{N} n_i\) is the total number of commanded actuator channels, with \(N\) the number of active modules and \(n_i\) the number of actuated joints in module \(i\). The checksum is evaluated over the packet body as
\begin{equation}
\chi_k=\mathcal{C}\!\left(\boldsymbol{\pi}^{\mathrm{body}}_k\right),
\label{eq:checksum_general}
\end{equation}
where \(\mathcal{C}(\cdot)\) denotes the protocol-specific checksum operator.

At each sampling instant, the master first requests the robot state and then transmits the updated actuator command. The returned measurement vector is written as
\begin{equation}
\mathbf{x}_k=
\begin{bmatrix}
\mathbf{q}_k^{\top} &
\dot{\mathbf{q}}_k^{\top} &
\mathbf{y}_{s,k}^{\top}
\end{bmatrix}^{\top},
\label{eq:measured_state_general}
\end{equation}
where
\begin{equation}
\mathbf{q}_k=
\begin{bmatrix}
\mathbf{q}_{1,k}^{\top} & \mathbf{q}_{2,k}^{\top} & \cdots & \mathbf{q}_{N,k}^{\top}
\end{bmatrix}^{\top},
\mathbf{q}_{i,k}=
\begin{bmatrix}
q_{1,i,k} & \cdots & q_{n_i,i,k}
\end{bmatrix}^{\top},
\end{equation}
and similarly for \(\dot{\mathbf{q}}_k\). Thus, in the present prototype, \(q_{1,1}\) and \(q_{2,1}\) denote the two actuator coordinates of the first module, while \(q_{1,2}\) and \(q_{2,2}\) denote those of the second module. The auxiliary sensor vector is
\begin{equation}
\mathbf{y}_{s,k}=
\begin{bmatrix}
\mathbf{y}_{s,1,k}^{\top} & \cdots & \mathbf{y}_{s,N,k}^{\top}
\end{bmatrix}^{\top},
\end{equation}
where \(\mathbf{y}_{s,i,k}\) may include IMU attitude and angular-rate measurements of module \(i\). All acquired signals are converted from native device units to SI units prior to control computation. A dedicated stop command is issued at termination to guarantee safe shutdown of the modular system.

\begin{table}[t]
\caption{Prototype, magnetic coupling, and control parameters.}
\label{tab:prototype_params}
\centering
\scriptsize
\setlength{\tabcolsep}{3pt}
\renewcommand{\arraystretch}{0.95}
\begin{tabular}{c p{2.5cm} c}
\hline
Symbol & Meaning & Value \\
\hline
\multicolumn{3}{c}{\textit{Magnetic coupling}} \\
\hline
\(D_c\) & Central magnet diameter & \(30\,\mathrm{mm}\) \\
\(D_s\) & Stabilizer magnet diameter & \(10\,\mathrm{mm}\) \\
\(R_c\) & Central magnet radius & \(15\,\mathrm{mm}\) \\
\(R_s\) & Stabilizer magnet radius & \(5\,\mathrm{mm}\) \\
\(h_m\) & Magnet thickness & \(20\,\mathrm{mm}\) \\
\(N_r\) & Number of stabilizer magnets & \(10\) \\
\(r_m\) & Ring pitch radius & \(\approx 22\,\mathrm{mm}\) \\
\(h_{\mathrm{PLA}}\) & Thickness of one PLA wall & \(3.2\,\mathrm{mm}\) \\
\(g_a\) & Remaining air gap & \(3.6\,\mathrm{mm}\) \\
\(g\) & Total magnetic stand-off & \(10\,\mathrm{mm}\) \\
\(B_r\) & N52 remanence used & \(1.45\,\mathrm{T}\) \\
\hline
\multicolumn{3}{c}{\textit{Communication and control}} \\
\hline
\(N\) & Active modules & \(2\) \\
\(n_i\) & Actuated joints per module & \(2\) \\
\(n\) & Total commanded channels & \(4\) \\
\(\mathbf{h}\) & Packet header & \([\texttt{0xAA},\texttt{0x55}]^\top\) \\
\((m_{\mathrm{vel}},m_{\mathrm{state}},m_{\mathrm{stop}})\) & Mode IDs & \((1,2,3)\) \\
\(T_s\) & Sampling period & \(0.01\,\mathrm{s}\) \\
\(\mathbf{K}_{p,i}\) & Per-module P gain matrix & \(\mathrm{diag}(1,1)\) \\
\(\mathbf{K}_{d,i}\) & Per-module D gain matrix & \(\mathrm{diag}(0.05,0.05)\) \\
\(\alpha\) & Velocity scale & \(0.02398\,\mathrm{rad\,s^{-1}}\)/count \\
\(\mathbf{D}_i\) & Per-module unit map & \(\alpha\mathbf{I}_{n_i}\) \\
\(\mathbf{u}_{\min,i}\) & Lower command bound & \(-150\) \\
\(\mathbf{u}_{\max,i}\) & Upper command bound & \(150\) \\
\(\rho\) & Sphere radius & \(0.102\,\mathrm{m}\) \\
\(\Delta\mathbf{q}_{i}\) & Example reference move & user-defined \\
\(T_f\) & Motion duration & user-defined \\
\hline
\end{tabular}
\end{table}

\subsection{Trajectory Generation and Tracking Control}
\label{sec:traj_control}

To generate smooth point-to-point motion for each actuated joint, the desired reference is described by a fifth-order polynomial, which is a standard choice when continuity of position, velocity, and acceleration is required at the endpoints \cite{lynch2017modern,siciliano2009robotics}. For the \(a\)-th actuator of module \(i\),
\begin{equation}
q_{d,a,i}(t)=\sum_{j=0}^{5} a_{j,a,i} t^{j}, 
\qquad t\in[0,T_f],
\label{eq:quintic_general}
\end{equation}
with corresponding velocity and acceleration
\begin{equation}
\dot q_{d,a,i}(t)=\sum_{j=1}^{5} j\,a_{j,a,i} t^{j-1},
\ddot q_{d,a,i}(t)=\sum_{j=2}^{5} j(j-1)\,a_{j,a,i} t^{j-2}.
\label{eq:quintic_derivatives}
\end{equation}
The polynomial coefficients are uniquely determined from the endpoint boundary conditions
\begin{equation*}
q_{d,a,i}(0)=q_{0,a,i}, \quad
\dot q_{d,a,i}(0)=\dot q_{0,a,i}, \quad
\ddot q_{d,a,i}(0)=\ddot q_{0,a,i},
\end{equation*}
\begin{equation}
q_{d,a,i}(T_f)=q_{f,a,i}, 
\dot q_{d,a,i}(T_f)=\dot q_{f,a,i}, \;
\ddot q_{d,a,i}(T_f)=\ddot q_{f,a,i}.
\end{equation}
These conditions give the linear system
\begin{equation}
\mathbf{M}(T_f)\mathbf{a}_{a,i}=\mathbf{b}_{a,i},
\label{eq:quintic_system}
\end{equation}
where
\begin{equation*}
\mathbf{a}_{a,i}=
\begin{bmatrix}
a_{0,a,i} & a_{1,a,i} & a_{2,a,i} & a_{3,a,i} & a_{4,a,i} & a_{5,a,i}
\end{bmatrix}^{\top},
\end{equation*}
\begin{equation}
\mathbf{b}_{a,i}=
\begin{bmatrix}
q_{0,a,i} & \dot q_{0,a,i} & \ddot q_{0,a,i} &
q_{f,a,i} & \dot q_{f,a,i} & \ddot q_{f,a,i}
\end{bmatrix}^{\top},
\end{equation}
and
\begin{equation}
\mathbf{M}(T_f)=
\begin{bmatrix}
1 & 0 & 0 & 0 & 0 & 0\\
0 & 1 & 0 & 0 & 0 & 0\\
0 & 0 & 2 & 0 & 0 & 0\\
1 & T_f & T_f^2 & T_f^3 & T_f^4 & T_f^5\\
0 & 1 & 2T_f & 3T_f^2 & 4T_f^3 & 5T_f^4\\
0 & 0 & 2 & 6T_f & 12T_f^2 & 20T_f^3
\end{bmatrix}.
\label{eq:quintic_matrix}
\end{equation}
Hence,
\begin{equation}
\mathbf{a}_{a,i}=\mathbf{M}(T_f)^{-1}\mathbf{b}_{a,i}.
\label{eq:quintic_solution}
\end{equation}

The quintic form is adopted because it guarantees continuity of position, velocity, and acceleration at both endpoints, thereby reducing abrupt transients in the commanded actuation \cite{lynch2017modern,siciliano2009robotics}. In the common rest-to-rest case,
\begin{equation}
\dot q_{0,a,i}=\dot q_{f,a,i}=0,
\qquad
\ddot q_{0,a,i}=\ddot q_{f,a,i}=0,
\label{eq:rest_to_rest}
\end{equation}
which yields a smooth start and stop for each actuator.

For module \(i\), the desired joint trajectory is assembled as
\begin{align}
\mathbf{q}_{d,i}(t)=
\begin{bmatrix}
q_{d,1,i}(t) & \cdots & q_{d,n_i,i}(t)
\end{bmatrix}^{\top},\nonumber\\
\dot{\mathbf{q}}_{d,i}(t)=
\begin{bmatrix}
\dot q_{d,1,i}(t) & \cdots & \dot q_{d,n_i,i}(t)
\end{bmatrix}^{\top}.
\label{eq:vector_reference_module}
\end{align}
Stacking all modules gives the full reference vector
\begin{align}
\mathbf{q}_{d}(t)=
\begin{bmatrix}
\mathbf{q}_{d,1}^{\top}(t) & \cdots & \mathbf{q}_{d,N}^{\top}(t)
\end{bmatrix}^{\top},\nonumber\\
\dot{\mathbf{q}}_{d}(t)=
\begin{bmatrix}
\dot{\mathbf{q}}_{d,1}^{\top}(t) & \cdots & \dot{\mathbf{q}}_{d,N}^{\top}(t)
\end{bmatrix}^{\top}.
\label{eq:vector_reference}
\end{align}
Thus, the variables plotted later in the experimental results, such as \(q_{1,1}\), \(q_{2,1}\), \(\dot q_{1,1}\), and \(\dot q_{2,1}\), correspond directly to the first and second actuator coordinates of module \(1\).

Using the measured actuator states, the tracking errors for module \(i\) are defined by
\begin{equation}
\mathbf{e}_{i}(t)=\mathbf{q}_{d,i}(t)-\mathbf{q}_{i}(t),
\qquad
\dot{\mathbf{e}}_{i}(t)=\dot{\mathbf{q}}_{d,i}(t)-\dot{\mathbf{q}}_{i}(t).
\label{eq:tracking_errors}
\end{equation}
The experimental tracking controller is selected as a diagonal joint-space PD law for each module:
\begin{equation}
\mathbf{u}_{\omega,i}(t)=\mathbf{K}_{p,i}\mathbf{e}_{i}(t)+\mathbf{K}_{d,i}\dot{\mathbf{e}}_{i}(t),
\label{eq:pd_law}
\end{equation}
where \(\mathbf{u}_{\omega,i}\in\mathbb{R}^{n_i}\) is the commanded actuator velocity of module \(i\) in SI units, and \(\mathbf{K}_{p,i},\mathbf{K}_{d,i}\in\mathbb{R}^{n_i\times n_i}\) are positive diagonal gain matrices \cite{lynch2017modern,siciliano2009robotics}.

Because the embedded motor interface accepts commands in native integer units, the control action is mapped through a conversion matrix and saturated before transmission:
\begin{equation}
\mathbf{u}_{i,k}=
\mathrm{sat}\!\left(
\mathrm{round}\!\left(\mathbf{D}_{i}^{-1}\mathbf{u}_{\omega,i,k}\right),
\mathbf{u}_{\min,i},
\mathbf{u}_{\max,i}
\right),
\label{eq:command_mapping}
\end{equation}
where \(\mathbf{D}_{i}\) is the unit-conversion matrix for module \(i\), and \(\mathbf{u}_{\min,i},\mathbf{u}_{\max,i}\in\mathbb{R}^{n_i}\) are the element-wise safety limits. In the current prototype, \(\mathbf{D}_{i}=\alpha\mathbf{I}_{n_i}\), with \(\alpha\) given in Table~\ref{tab:prototype_params}. During each experimental run, the controller records the measured states, reference trajectories, tracking errors, transmitted motor commands, and IMU outputs for subsequent analysis. The present implementation focuses on actuator-level tracking, where desired rolling behaviours are generated through prescribed pendulum joint references. Extension to robot-level motion tracking, such as desired planar \(x\)-\(y\) trajectories of the rolling shell, would require an additional mapping from task-space motion objectives to internal pendulum references. Such a mapping is an important next step, but lies beyond the scope of the current experimental study.

\subsection{IMU-Based Planar Motion Reconstruction}

Beyond actuator-level tracking, the recorded IMU data are used in post-processing to estimate the planar motion of each rolling shell. Let \(\mathbf{R}_{i,k}\in SO(3)\) denote the orientation matrix of module \(i\) at sample \(k\). In this work, a \(ZYX\) Euler-angle convention is adopted, such that
\begin{equation}
\mathbf{R}_{i,k}
=
\mathbf{R}_z(\theta_{z,i,k})
\mathbf{R}_y(\theta_{y,i,k})
\mathbf{R}_x(\theta_{x,i,k}),
\label{eq:euler_zyx}
\end{equation}
where \(\theta_{x,i,k}\), \(\theta_{y,i,k}\), and \(\theta_{z,i,k}\) denote the roll, pitch, and yaw channels, respectively. In the present notation, these channels correspond to the measured orientation variables \(\theta_{x,i,k}\), \(\theta_{y,i,k}\), and \(\theta_{z,i,k}\), as plotted for example in the results section for module \(1\). Assuming pure rolling on a locally planar surface, the body-fixed contact direction is written as
\begin{equation}
\mathbf{c}_{i,k}=\rho\,\mathbf{R}_{i,k}^{\top}\mathbf{e}_3^{-},
\qquad
\mathbf{e}_3^{-}=
\begin{bmatrix}
0 & 0 & -1
\end{bmatrix}^{\top},
\label{eq:contact_direction}
\end{equation}
where \(\rho\) is the effective sphere radius.

The incremental orientation change between two successive samples is computed as
\begin{equation}
\Delta\mathbf{R}_{i,k}=\mathbf{R}_{i,k}\mathbf{R}_{i,k-1}^{\top}.
\label{eq:R_increment}
\end{equation}
Using the matrix logarithm, the corresponding incremental rotation vector is
\begin{equation}
\boldsymbol{\phi}_{i,k}=
\mathrm{Log}\!\left(\Delta\mathbf{R}_{i,k}\right)^{\vee},
\label{eq:rotation_vector}
\end{equation}
where \((\cdot)^{\vee}\) denotes the standard mapping from \(\mathfrak{so}(3)\) to \(\mathbb{R}^{3}\), as commonly used in Lie-group formulations of rigid-body kinematics \cite{lynch2017modern}. Under the no-slip approximation, the induced planar displacement increment is
\begin{equation}
\Delta\mathbf{p}_{i,k}=
\rho
\begin{bmatrix}
0 & 1 & 0\\
-1 & 0 & 0
\end{bmatrix}
\boldsymbol{\phi}_{i,k},
\label{eq:planar_increment}
\end{equation}
and the horizontal position estimate is recovered recursively as
\begin{equation}
\mathbf{p}_{i,k}=\mathbf{p}_{i,k-1}+\Delta\mathbf{p}_{i,k},
\qquad
\mathbf{p}_{i,k}\in\mathbb{R}^{2}.
\label{eq:planar_position}
\end{equation}

This reconstruction yields an approximate \(XY\) trajectory of each rolling module and is used only for motion interpretation and experimental post-analysis. It does not affect the real-time control loop.

\section{Motion Analysis}
\label{sec:results}

Experimental validation was carried out using two wireless spherical modules instrumented with onboard IMUs and joint-state feedback. The reported experiments were selected to assess four representative behaviours: single-module circular rolling, approach-and-couple interaction, relative spinning between coupled modules, and joint spinning with maintained contact. For each case, measured joint trajectories and module orientations are reported together with representative snapshots of the physical motion.

\begin{figure}[t!]
\centering
\includegraphics[width= 3.6 in]{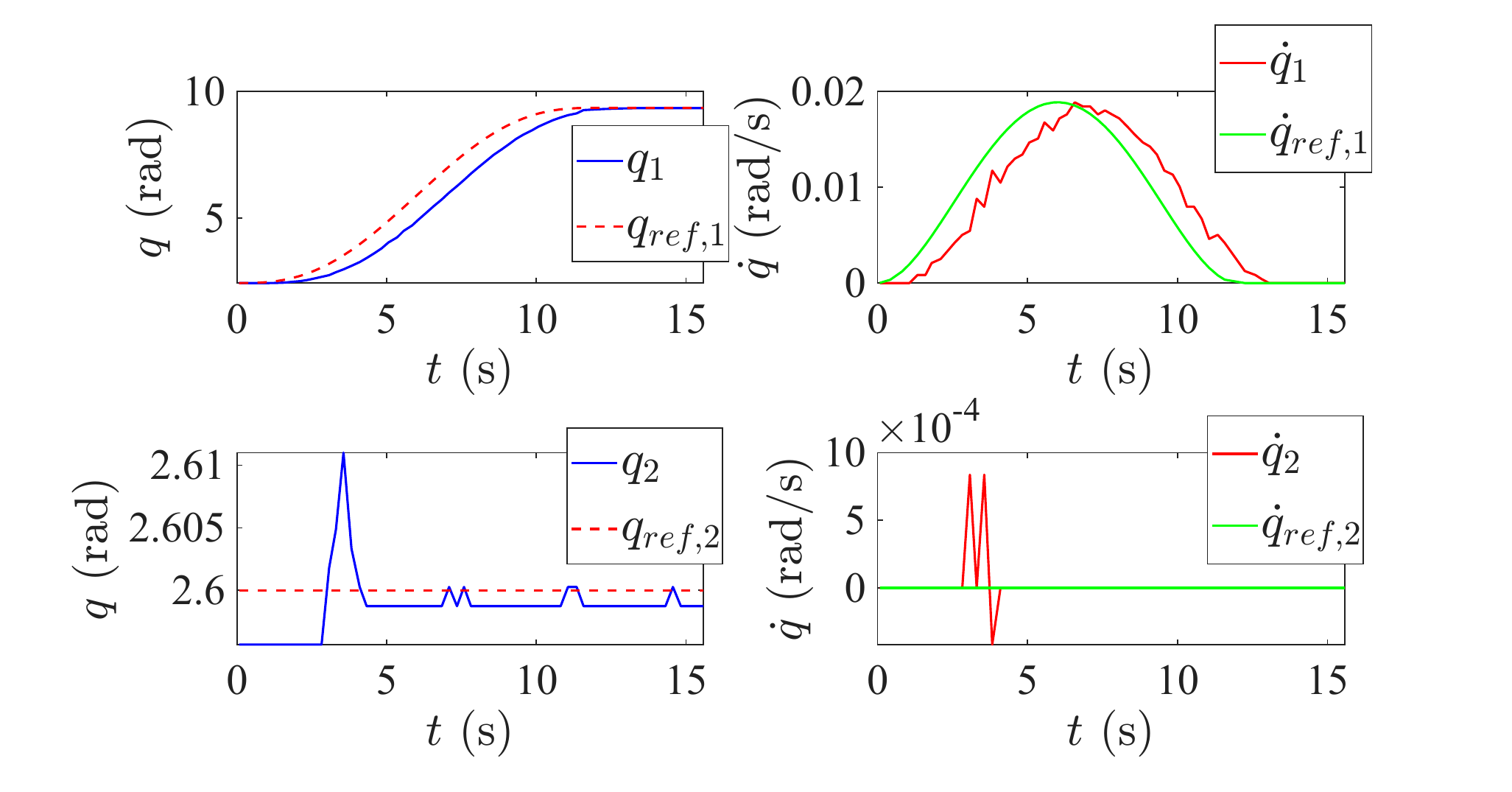}
\includegraphics[width= 3.6 in]{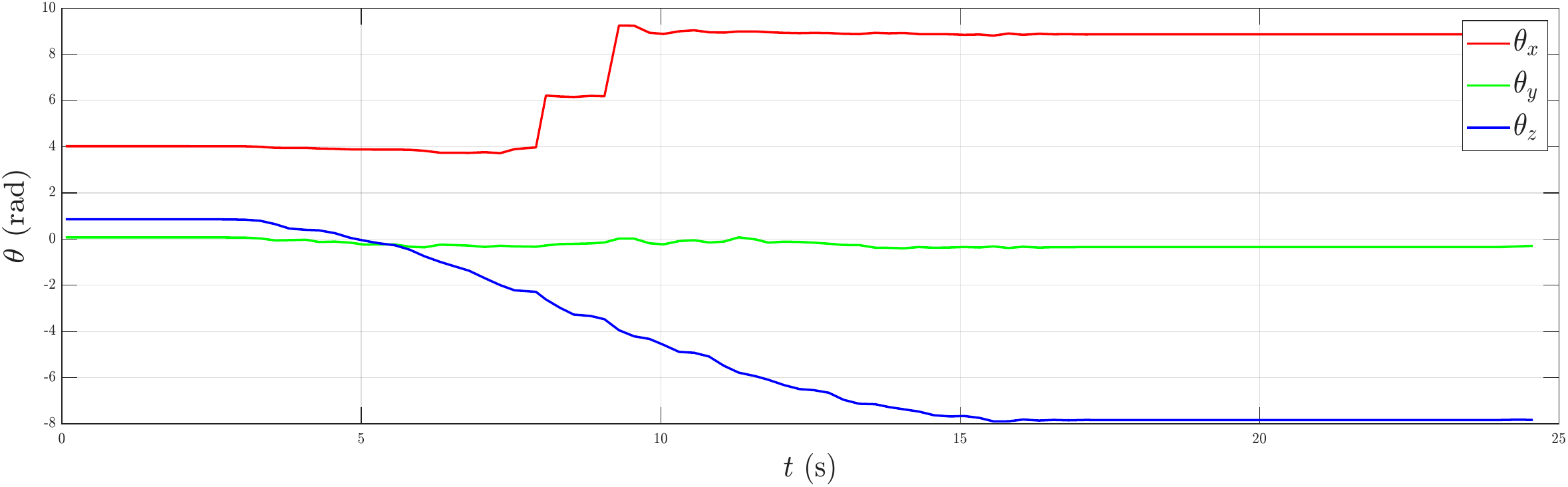}
\includegraphics[width=\linewidth]{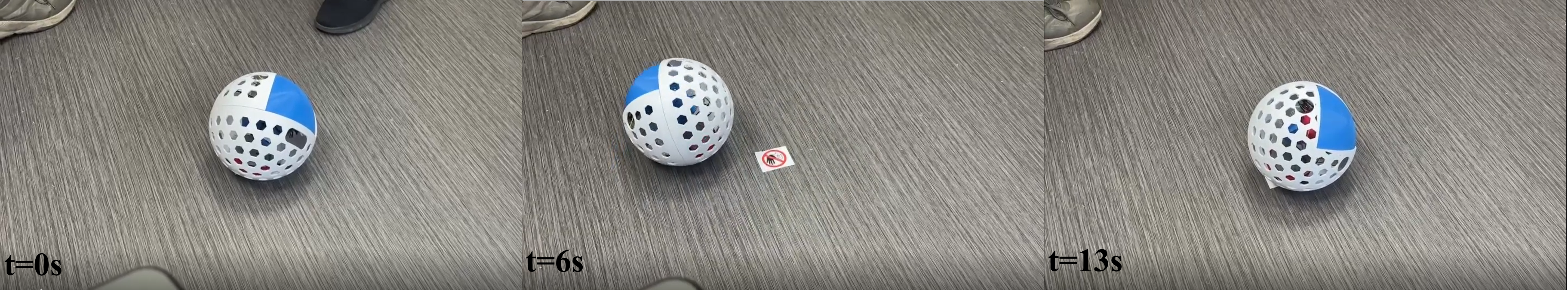}
\caption{Single-module circular rolling: measured and reference joint states, module orientation, and motion snapshots.}
\label{fig:circle1}
\end{figure}

Fig.~\ref{fig:circle1} shows the baseline circular-rolling response of a single module under the reference generation and joint-space PD tracking law in \eqref{eq:vector_reference}--\eqref{eq:pd_law}. The primary actuated coordinate \(q_{1,1}\) follows its quintic reference with a moderate transient lag, while remaining stable and bounded throughout the manoeuvre. In contrast, \(q_{2,1}\), which is regulated to a nearly constant target through the same error dynamics in \eqref{eq:tracking_errors}, exhibits only small deviations about its prescribed value. The corresponding velocity responses also remain bounded, indicating that the controller in \eqref{eq:pd_law}, together with the command mapping in \eqref{eq:command_mapping}, is sufficient to produce repeatable curved rolling motion at the module level.

The IMU signals further confirm that the circular manoeuvre is achieved through a coordinated reorientation of the shell rather than through erratic body oscillation. In particular, the measured attitude evolves smoothly over the manoeuvre and the image sequence shows a consistent arc-like displacement. This experiment therefore establishes the nominal single-module tracking fidelity used as the reference case for subsequent coupled-motion tests.

\begin{figure}[t!]
\centering
\includegraphics[width=\linewidth]{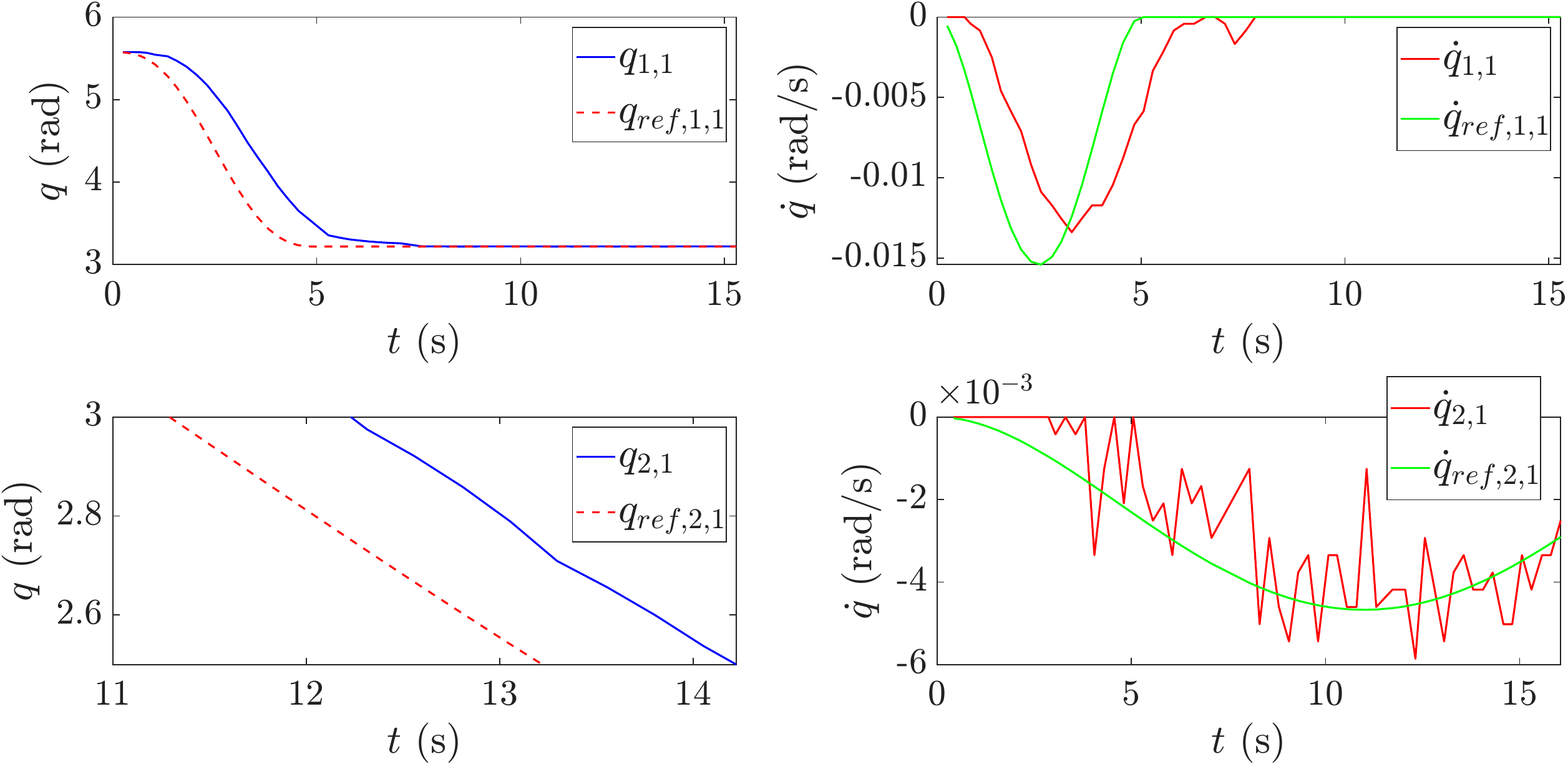}
\includegraphics[width=\linewidth]{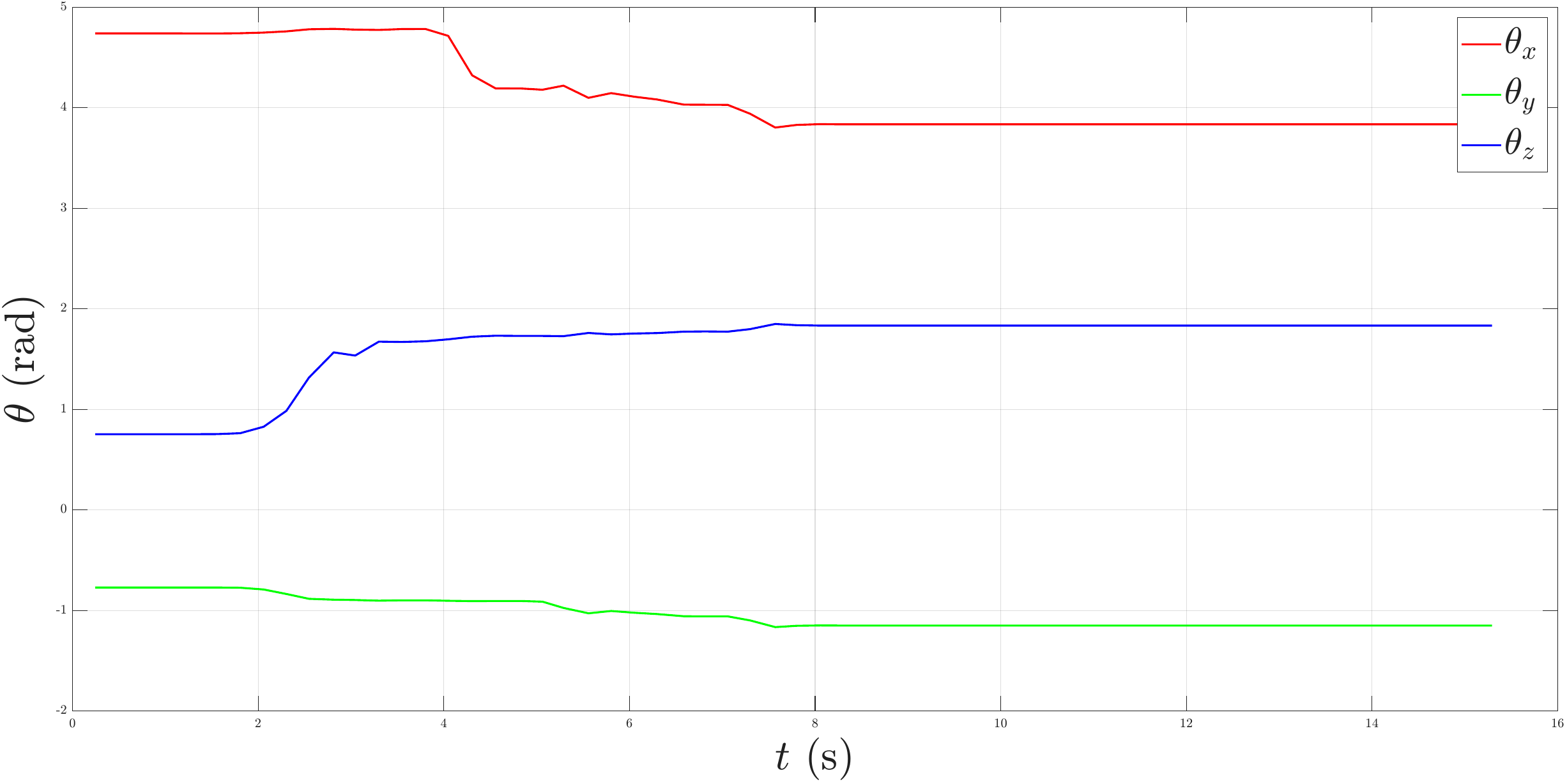}
\includegraphics[width=\linewidth]{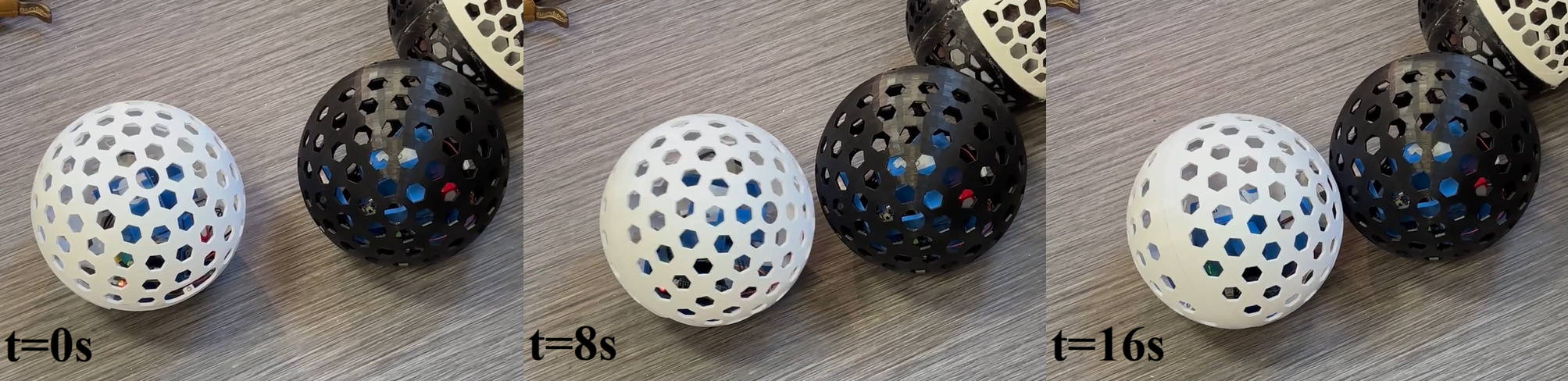}
\caption{Coupling process between two robots measured via onboard IMU.}
\label{fig:coupling}
\end{figure}

Fig.~\ref{fig:coupling} illustrates the transition from independent rolling to physical docking. During the approach phase, the measured orientation of the active module evolves smoothly toward a new equilibrium after contact, with no evidence of large overshoot or unstable rebound. The joint response remains bounded throughout the manoeuvre, indicating that the docking event does not destabilize the actuator-level control loop.

The snapshots confirm successful mechanical capture and subsequent retention of the coupled state. Together with the bounded attitude response, this indicates that the proposed magnetic interface provides sufficient passive alignment and retention to support repeatable module-to-module attachment under rolling interaction.

\begin{figure}[t!]
\centering
\includegraphics[width=\linewidth]{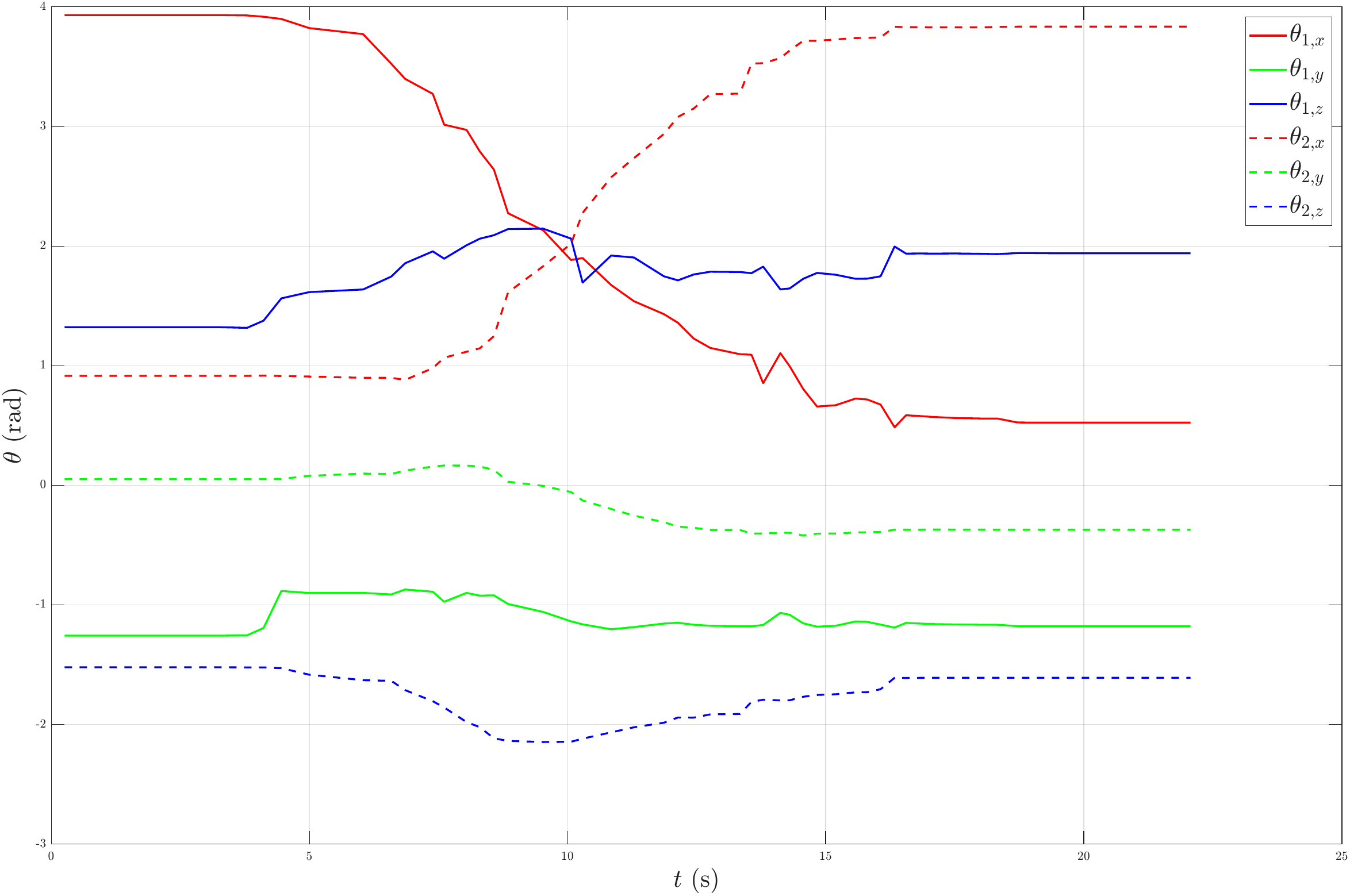}
\includegraphics[width=\linewidth]{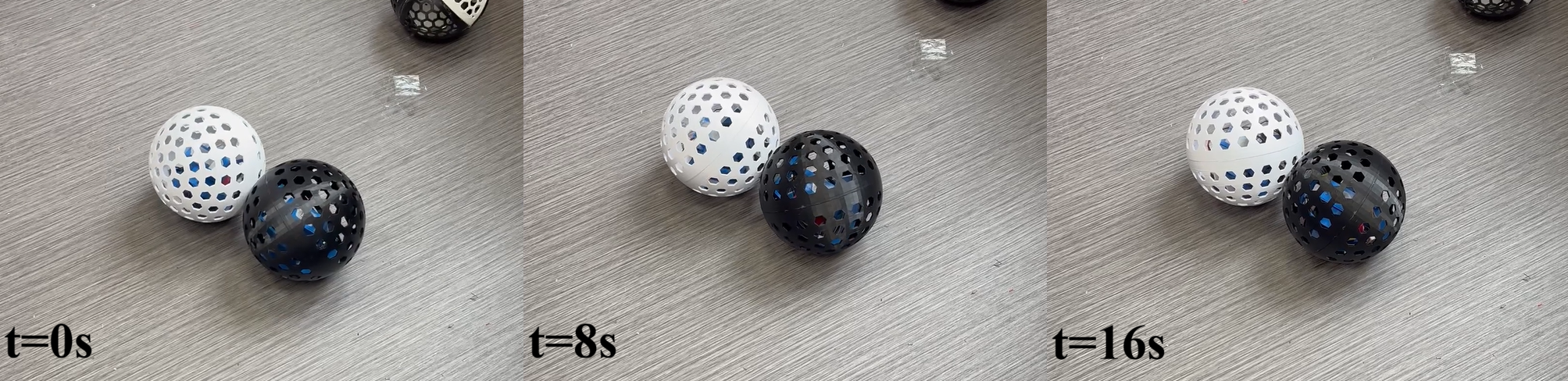}
\caption{Relative spinning process between two robots measured via onboard IMU.}
\label{fig:spinning}
\end{figure}

\begin{figure}[t!]
\centering
\includegraphics[width=\linewidth]{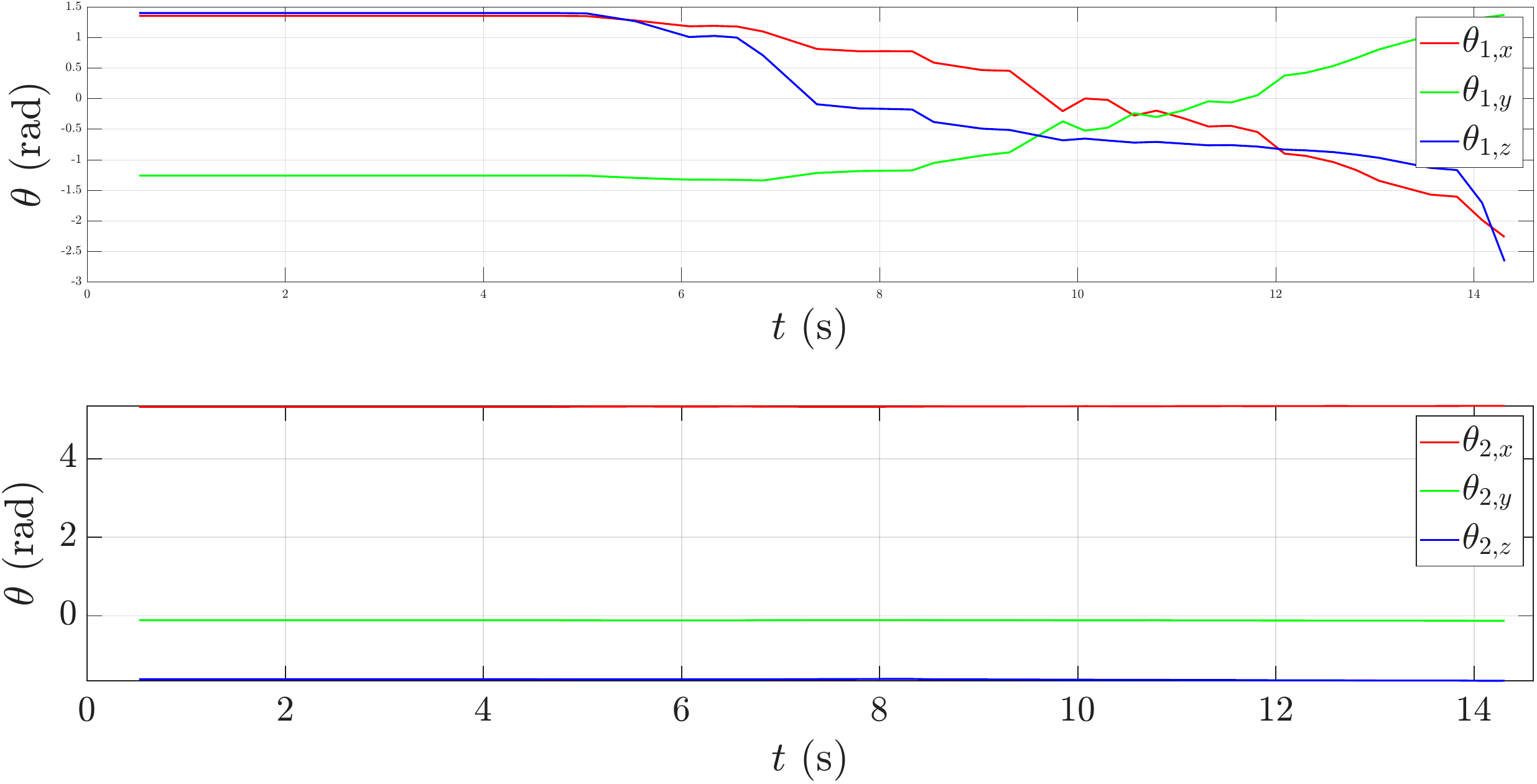}
\includegraphics[width=\linewidth]{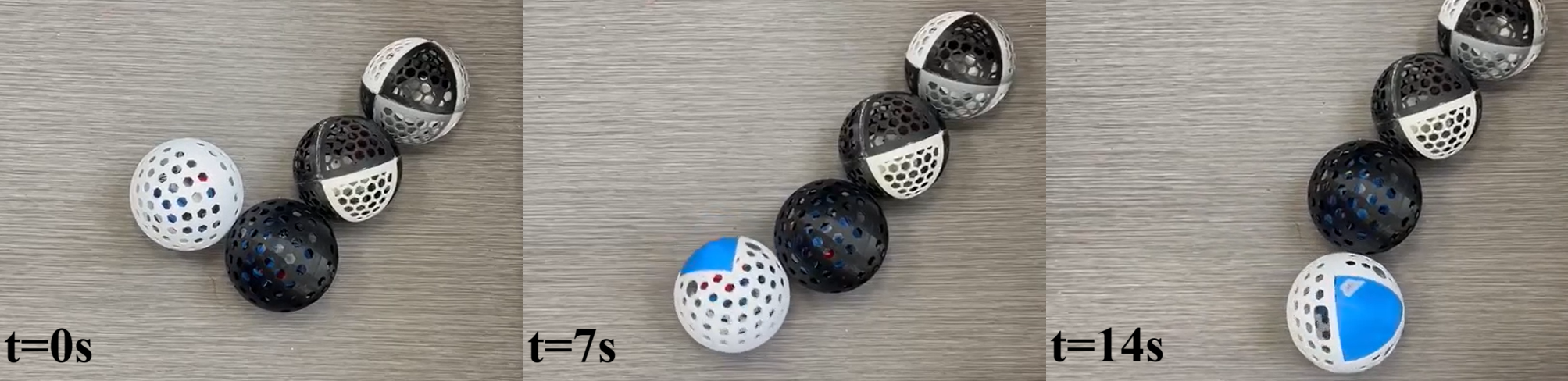}
\caption{Joint spinning motion with maintained contact between two robots.}
\label{fig:joint}
\end{figure}

Figs.~\ref{fig:spinning} and~\ref{fig:joint} show two distinct coupled-motion regimes generated by different actuation patterns. In Fig.~\ref{fig:spinning}, both motors of each spherical module are commanded so that the coupled bodies generate relative spinning while remaining in contact. This produces a motion in which the two spheres rotate against one another about the coupling interface rather than behaving as a single rigid pair. The measured orientations of both modules remain bounded, confirming that the coupled system does not diverge during interaction. However, compared with the single-module case, the response is visibly noisier, especially in the actuator velocity signals. This suggests intermittent stick--slip behaviour and local contact disturbances at the shell--shell and magnetic interface, likely associated with repeated micro-separation, frictional re-engagement, and small impact-like events during relative rotation.

By contrast, Fig.~\ref{fig:joint} corresponds to a joint-motion regime in which the base-side actuation of one module drives the coupled end-effector interaction so that the two spheres move in a more cooperative manner. In this case, the magnetic contact transmits the motion more like a constrained jointed pair, and the second sphere follows the imposed motion rather than spinning freely relative to the first. The measured attitude traces remain smoother than in the relative-spinning case, and no large drift in the coupled configuration is observed. This indicates that the base-driven coupled mode generates a more stable joint-like behaviour, whereas simultaneous multi-motor actuation of both modules tends to promote relative spinning and stronger contact-induced disturbances.

\section{Conclusion}

This paper presented \emph{PenduMorph}, a wireless pendulum-actuated rolling spherical robot with integrated magnetic docking for reconfigurable modular interaction. The platform combines enclosed rolling locomotion, onboard actuation and sensing, and a passive magnetic interface that enables both independent and coupled behaviours. An analytical model of the magnetic coupling was developed to estimate retention capability and to clarify the respective roles of the central and stabilizer magnets. Experimental results demonstrated stable single-module rolling, repeatable magnetic coupling, relative spinning between coupled modules, and joint spinning with maintained contact. While the relative-spinning case exhibited increased oscillation and velocity noise, the coupled configuration remained bounded and mechanically robust throughout the tested interactions. These results indicate that the proposed platform provides a useful experimental foundation for future reconfigurable spherical rolling robots. Future work will focus on richer coupled-motion control, improved contact-state estimation, and extension to larger multi-module assemblies.








 \bibliographystyle{ieeetr}
 \bibliography{references}

  \end{document}